\documentclass[letterpaper, 10 pt, conference]{ieeeconf}  
\usepackage[utf8]{inputenc}
\usepackage[T1]{fontenc}
\usepackage{hyperref}
\usepackage{url}
\usepackage{booktabs}
\usepackage{multirow}
\usepackage{graphicx}
\usepackage{xspace}
\usepackage{amsmath}
\usepackage{amssymb}
\usepackage{cuted}
\usepackage{subcaption}
\usepackage{xcolor}
\usepackage{wrapfig}
\usepackage{cite}

\definecolor{DeepPink}{RGB}{255,20,147}

\IEEEoverridecommandlockouts                              

\title{\LARGE \bf
UltraDexGrasp: Learning Universal Dexterous Grasping for \\
Bimanual Robots with Synthetic Data
}

\author{Sizhe Yang$^{1,2}$\quad Yiman Xie$^{1,3}$\quad Zhixuan Liang$^{1,4}$\quad Yang Tian$^{1,5}$\quad Jia Zeng$^{1}$ \\
Dahua Lin$^{1,2}$\quad Jiangmiao Pang$^{1}$ \\ 
$^{1}$Shanghai AI Laboratory 
\quad $^{2}$The Chinese University of Hong Kong \\
$^3$Zhejiang University \quad $^4$The University of Hong Kong \quad $^5$Peking University\\
Project page: \textcolor{DeepPink}{\url{https://yangsizhe.github.io/ultradexgrasp/}
}
}

\begin{document}

\maketitle
\thispagestyle{empty}
\pagestyle{empty}

\begin{strip}
\centering
\vspace{-8mm}
\includegraphics[width=\textwidth]{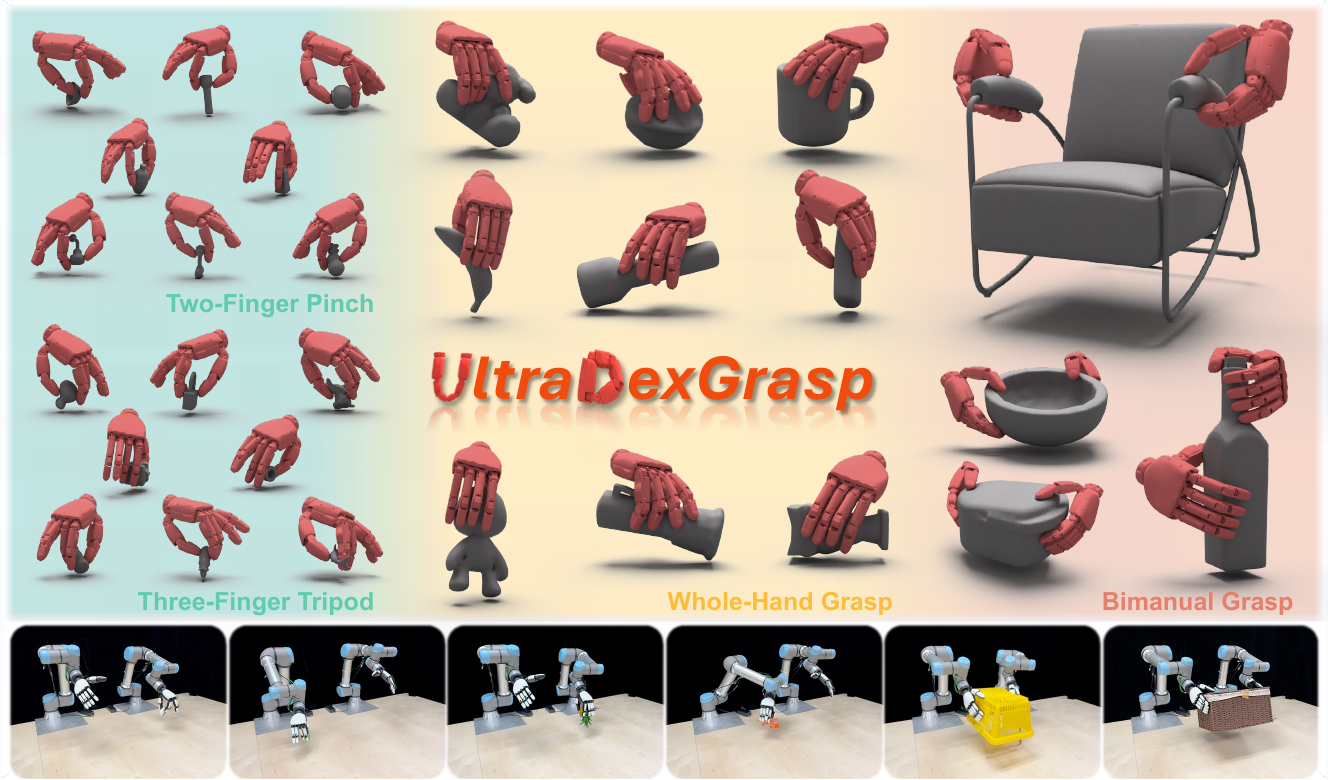}
\vspace{-3mm}
\captionof{figure}{\textbf{Overview.} \textbf{UltraDexGrasp} is a framework for universal dexterous grasping with bimanual robots. The proposed data generation pipeline integrates an optimization-based grasp synthesizer with a planning-based demonstration generation module, and supports multiple grasp strategies, including two-finger pinch, three-finger tripod, whole-hand grasp, and bimanual grasp. Trained on data produced by this pipeline, the policy demonstrates robust zero-shot sim-to-real transfer and strong generalization to novel objects with varied shapes, sizes, and weights.}
\label{fig:teaser}
\end{strip}

\begin{abstract}
Grasping is a fundamental capability for robots to interact with the physical world. 
Humans, equipped with two hands, autonomously select appropriate grasp strategies based on the shape, size, and weight of objects, enabling robust grasping and subsequent manipulation. In contrast, current robotic grasping remains limited, particularly in multi-strategy settings.
Although substantial efforts have targeted parallel-gripper and single-hand grasping, dexterous grasping for bimanual robots remains underexplored, with data being a primary bottleneck.
Achieving physically plausible and geometrically conforming grasps that can withstand external wrenches poses significant challenges.
To address these issues, we introduce \textit{UltraDexGrasp}, a framework for universal dexterous grasping with bimanual robots. 
The proposed data-generation pipeline integrates optimization-based grasp synthesis with planning-based demonstration generation, yielding high-quality and diverse trajectories across multiple grasp strategies. With this framework, we curate \textit{UltraDexGrasp-20M}, a large-scale, multi-strategy grasp dataset comprising 20 million frames across 1,000 objects.
Based on UltraDexGrasp-20M, we further develop a simple yet effective grasp policy that takes point clouds as input, aggregates scene features via unidirectional attention, and predicts control commands.
Trained exclusively on synthetic data, the policy achieves robust zero-shot sim-to-real transfer and consistently succeeds on novel objects with varied shapes, sizes, and weights, attaining an average success rate of 81.2\% in real-world universal dexterous grasping. To facilitate future research on grasping with bimanual robots, we open-source the data generation pipeline at \href{https://github.com/InternRobotics/UltraDexGrasp}{\textcolor{DeepPink}{https://github.com/InternRobotics/UltraDexGrasp}}.
\end{abstract}

\section{Introduction}
\label{sec:introduction}

Dexterous grasping is a critical step toward enabling robots to manipulate objects with human-level proficiency. As the first stage of many manipulation skills, it bridges the transition from non-contact to contact interaction.
In daily life, humans autonomously select grasping strategies that accommodate variations in the shape, size, and weight of objects: large, heavy objects generally require a coordinated bimanual grasp to maintain balance;
medium-sized objects can be stably grasped by a single hand using all five fingers; and small objects, which provide insufficient surface area, are best handled with two-finger pinch or three-finger tripod. Humans can also adapt hand postures to object geometry, maximizing surface conformity to ensure robust grasping and subsequent manipulation.
In contrast, current robotic grasping capabilities fall far short of human performance. Although the burgeoning humanoid-robot market has accelerated the development of dual-arm platforms and dexterous hands, data and algorithms remain the principal bottleneck.
Considerable efforts~\cite{anygrasp, anydexgrasp, dexgraspnet, dexgraspvla, graspvla, graspnet, drograsp}
have been devoted to parallel-gripper and single-hand grasping, yet universal dexterous grasping for bimanual robots remains underexplored.

A central difficulty of universal dexterous grasping is producing high-quality data.
Existing data generation paradigms for dexterous grasping fall into three strands:
1) reinforcement learning (RL) with privileged information that trains experts for grasping~\cite{unidexgrasp++, dextrahg, dextrahrgb, dexhanddiff}; 
2) optimization-based synthesis that minimizes grasp-related costs~\cite{dexgraspnet, liu2021synthesizing, li2023frogger, turpin2022grasp}; 
and 3) learning-based synthesis that trains generative models to predict diverse grasps~\cite{unidexgrasp, dex1b, jiang2021hand}. 
However, RL-trained experts are typically deterministic after training, thus mapping a given observation to a single action, which leads to repetitive postures and a lack of diversity.
Optimization- and learning-based synthesis are largely open-loop, struggle in dynamic real-world scenarios and typically ignore arm kinematics. Moreover, previous approaches are predominantly limited to single-hand settings.
Universal dexterous grasping for dual-arm robots poses exceptional challenges for data generation, due to the extensive degrees of freedom, requirements for bimanual coordination, and the multitude of possible grasping strategies.
Whether data are collected via real-world teleoperation or generated in simulation, producing physically plausible and geometrically conforming grasps that resist external wrenches, is essential though difficult.

To address these challenges, we propose \textbf{UltraDexGrasp}, a framework for universal dexterous grasping with bimanual robots. The proposed data-generation pipeline integrates an optimization-based grasp synthesizer with a planning-based demonstration generation module for coordinated dual-arm manipulation. 
This integration yields kinematically feasible and natural closed-loop motions while preserving data diversity.
Notably, the data generation supports multiple grasp strategies, including holding large objects with both hands, whole-hand grasp for medium-sized objects, and two-finger pinch or three-finger tripod for small objects.
With this pipeline, we curate \textbf{UltraDexGrasp-20M}, the first large-scale multi-strategy dexterous grasp dataset for bimanual robots, comprising 20 million frames over 1,000 objects. 

Building upon this, we develop a simple yet effective grasp policy that takes point clouds as input, aggregates scene features via unidirectional attention, and predicts control commands, enabling multiple grasp strategies and improving generalization across diverse objects. We conduct comprehensive experiments in both simulation and the real world to evaluate the robustness of the proposed policy trained on UltraDexGrasp-20M. Testing in simulation on 600 objects---including both seen and unseen objects during training---that vary widely in shape, weight (ranging from 5~g to 1,000~g), and size (from objects with a longest bounding box edge less than 0.03~m to those with a shortest edge exceeding 0.5~m), our policy attains an 84.0\% average success rate, exceeding the next best baseline by 25.2 percentage points (approximately 43\% relative improvement).
The policy, trained exclusively on synthetic data, is further deployed in real-world scenarios. It adapts grasp strategies to different objects and successfully handles diverse items, achieving an 81.2\% success rate across all objects.

In summary, our paper makes three contributions. First, we present UltraDexGrasp-20M and its data-generation pipeline that integrates optimization-based grasp synthesis with planning-based demonstration generation, producing high-quality and diverse trajectories across multiple grasp strategies. Second, we introduce a novel grasp policy for universal dexterous grasping with bimanual robots that enhances generalization to diverse objects. Third, we demonstrate that our policy, trained solely on synthetic data, enables robust zero-shot sim-to-real transfer and strong generalization to novel objects with varied shapes, sizes, and weights.

\section{Related Work}
\label{sec:related_work}

\textbf{Dexterous grasp synthesis and dataset.}
Grasp synthesis is crucial for the advancement of dexterous grasping, and existing approaches can be broadly categorized into three types: sampling-based, optimization-based, and learning-based methods. Sampling-based methods~\cite{graspit} typically require simplification of the search space, which leads to limited grasp diversity. Many studies focus on optimization-based methods~\cite{dexgraspnet, liu2021synthesizing, li2023frogger, turpin2022grasp, turpin2023fast, graspqp}, with some utilizing differentiable force closure to optimize grasp poses~\cite{dexgraspnet, liu2021synthesizing, li2022gendexgrasp}. \cite{bodex} formulates grasp synthesis as a bilevel optimization problem to generate high-quality grasps. Supervised learning-based methods~\cite{unidexgrasp, dex1b, wei2022dvgg, jiang2021hand} require a small amount of grasp data to train generative models that can produce a large number of grasps for novel objects. Additionally, some works employ reinforcement learning to train experts for grasp generation~\cite{unidexgrasp++, dextrahg, dextrahrgb}. Effective grasp synthesis and teleoperation have enabled the creation of several dexterous grasp datasets~\cite{goldfeder2009columbia, dex1b, contactpose, dexycb, realdex}. 
Recently, \cite{bimangrasp} proposed a bimanual grasp generation algorithm, but it imposes a simplified
contact model, does not consider dual-arm coordination, and has not explored closed-loop control policies for universal  grasping. Although \cite{lin2025sim} explores bimanual dexterous grasping, the RL-trained expert can only grasp limited objects, such as boxes. Scaling to a wider variety of objects is costly and generalization remains challenging. 
In contrast, our work proposes a framework for generating grasps for arbitrary objects and trains policies capable of handling novel and diverse objects.

\textbf{Generalizable policy for robotic grasping.}
Grasping is a crucial skill for robotic manipulation. Several studies focus on data-efficient learning for grasping using real-world datasets~\cite{anygrasp, graspnet, 6dofgraspnet, dexgraspvla}. Some other works train and evaluate grasping policies in simulated environments, demonstrating high grasp success rates across large numbers of objects~\cite{unidexgrasp, unidexgrasp++, dexgraspanything}. Some research targets functional grasping, emphasizing object part functionality and subsequent tasks~\cite{fungrasp, dexvlg}. Recently, sim-to-real transfer of grasping policies has shown promising results. Methods such as \cite{graspvla, dexgraspnet2.0, getagrip, dex1b} generate large-scale grasping datasets using optimization-based or rule-based pipelines. Other approaches, including \cite{dextrahg, dextrahrgb, robustdexgrasp, lin2025sim, clutterdexgrasp}, employ reinforcement learning to train expert agents for grasp data generation, achieving strong real-world performance. Our proposed bimanual grasp policy is trained on data produced through the complementary integration of optimization-based and rule-based approaches, achieving robust real-world bimanual dexterous grasping.

\section{Preliminaries}
\label{sec:preliminaries}

\subsection{Definition of Grasp Pose for Bimanual Robots}

A grasp is defined as the act of restraining the motion of an object by applying forces and torques at a set of contact points. The grasp pose represents the critical configuration of the hands required to achieve a stable grasp. Specifically, we parameterize the grasp pose for bimanual robots as a tuple:
\begin{equation}
    g = \left\{ \left( \boldsymbol{t}_h, \boldsymbol{R}_h, \boldsymbol{q}_h \right) \mid h = 0, 1 \right\}, 
\end{equation}
where $h \in \{0, 1\}$ denotes the index of the hand, $\boldsymbol{t}_h \in \mathbb{R}^3$ is the translation of the $h$-th hand, $\boldsymbol{R}_h \in SO(3)$ represents the rotation of the hand, and $\boldsymbol{q}_h \in \mathbb{R}^n$ denotes joint positions for the hand, with $n$ being the number of joints of the hand.

\subsection{Basics of Grasp Modeling}

Grasping focuses on restraining the motion of an object through application of forces and torques at a set of contact points. We use point-on-plane contacts as the contact type, and hard finger model as the contact model, which is commonly used in force closure grasps. In the hard finger model, forces of arbitrary magnitude and direction can be applied within the limits defined by the friction cone:
\begin{equation}
    \mathcal{F} = \left\{ \boldsymbol{f} \;\middle|\; \|\boldsymbol{f}_{\mathrm{tan}}\| \leq \mu \|\boldsymbol{f}_{\mathrm{n}}\|,\;\; f_z \geq 0 \right\}, 
\end{equation}
where $\mu$ is the static friction coefficient of the object, 
$\boldsymbol{f}_{\mathrm{n}} = [0, 0, f_\mathrm{z}]$ is the vector component along the normal direction, and
$\boldsymbol{f}_{\mathrm{tan}} = [f_\mathrm{x}, f_\mathrm{y}, 0]$ is the vector component tangent to the surface.
Each contact point applies a wrench to the manipulated object:
\begin{equation}
    \boldsymbol{w}_i = \begin{bmatrix}
        \boldsymbol{f}_i \\
        \alpha (\boldsymbol{d}_i \times \boldsymbol{f}_i)
    \end{bmatrix}, 
\end{equation}
where $\boldsymbol{f}_i \in \mathbb{R}^3$ is the force, $\alpha(\boldsymbol{d}_i \times \boldsymbol{f}_i)$ is the torque, $\alpha \in \mathbb{R}$ is an arbitrary constant, and $\boldsymbol{d}_i$ is the relative position of the $i$-th contact point with respect to the center of mass of the object. $\boldsymbol{w}_i$ can be represented as $\boldsymbol{G}_i \boldsymbol{f}_i$, where $\boldsymbol{G}_i$ is a wrench basis matrix for the $i$-th contact. For hard finger model, the grasp map $\boldsymbol{G}_i$ is defined as:
\begin{equation}
  \boldsymbol{G}_i = 
    \begin{bmatrix}
      \mathbf{I}_{3\times 3} \\
      (\boldsymbol{p}_i - \boldsymbol{m})_{\times}
    \end{bmatrix}
  \boldsymbol{O}_i,  
\end{equation}
where $\mathbf{I}_{3 \times 3}$ is the $3 \times 3$ identity matrix, $(\boldsymbol{p}_i - \boldsymbol{m})_{\times}$ represents the cross product with $(\boldsymbol{p}_i - \boldsymbol{m})$, where $\boldsymbol{p}_i$ denotes the contact position, $\boldsymbol{m}$ represents the center of mass of the object, and $\boldsymbol{O}_i$ is the rotation matrix from the local contact frame to the frame of the object.

Thus, the grasp wrench space $\mathcal{W}$ for a grasp with $k$ contact points is defined as the set of possible wrenches $w$ that can be applied to the object:
\begin{equation}
    \mathcal{W} := \left\{ \boldsymbol{w} \;\middle|\; \boldsymbol{w} = \sum_{i=1}^k \boldsymbol{G}_i \boldsymbol{f}_i,~ \boldsymbol{f}_i \in \mathcal{F}_i,\;\; i=1,\ldots,k \right\}. 
\end{equation}

The grasp wrench space $\mathcal{W}$ is expected to be sufficiently large to resist external wrenches.

\section{Universal Dexterous Grasp Dataset}
\label{sec:dataset}

\begin{figure*}[ht]
    \centering
    \includegraphics[width=0.98\linewidth]{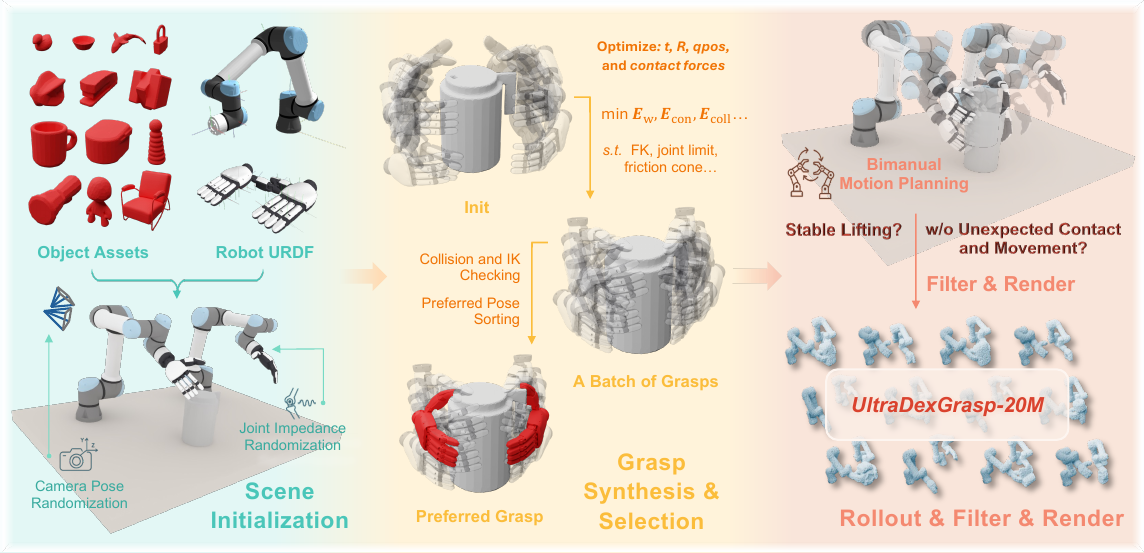}
    \caption{\textbf{Overview of data generation pipeline.} We first collect diverse object assets and import the objects and the robot URDF files into the simulator. An optimization-based grasp synthesizer is then used to generate feasible grasps, from which the preferred grasp is selected. Finally, motion planning is employed to generate demonstration trajectories.}
    \label{fig:data_pipeline}
    \vspace{-3mm}
\end{figure*}

As illustrated in Fig.~\ref{fig:data_pipeline}, the pipeline begins with scene initialization, where object assets and the robot URDF are imported into the simulation environment. 
We select 1,000 distinct objects from DexGraspNet~\cite{dexgraspnet} to construct our object assets.
Camera pose and joint impedance are randomized to reduce the sim-to-real gap. Given the meshes and poses of both the table and object, the grasp synthesizer generates a batch of feasible grasps, which are subsequently filtered and ranked to select the best candidate. We employ bimanual motion planning to compute collision-free and coordinated trajectories, which are then executed in simulation. Utilizing this scalable data generation pipeline, we curate the UltraDexGrasp-20M dataset.

\subsection{Grasp Synthesis}
\label{sec:grasp_synthesis}

Grasp synthesis involves two main stages. First, the hands are initialized near the object, and then, through optimization, physically plausible and geometrically conforming grasps are generated. In the initialization stage, we first obtain the convex hull of the object's mesh. 
For unimanual grasps, such as whole-hand, two-finger pinch, and three-finger tripod, we sample a point on the surface of the object’s convex hull. For bimanual grasps, two points are sampled on the convex hull surface, positioned on opposite sides of the object's center. The hand is positioned along the normal vector at the sampled point, with the palm facing the target object. 

Given the object mesh $\boldsymbol{M}$ with center of mass $\boldsymbol{m}$, two hands $h \in \{0,1\}$ with known kinematics, candidate contact sets $\bar{\mathcal{C}}_h$, a strategy-selected active contact set $\mathcal{C} \subset \bar{\mathcal{C}}_0 \cup \bar{\mathcal{C}}_1$, and the contact point $\boldsymbol{c}\in\mathcal{C}$, the decision variables are the bimanual grasp pose $\boldsymbol{g} = \{ (\boldsymbol{t}_h, \boldsymbol{R}_h, \boldsymbol{q}_h) \mid h = 0,1 \}$ and the contact forces $\{ \boldsymbol{f}_{\boldsymbol{c}} \in \mathbb{R}^3 \}_{\boldsymbol{c} \in \mathcal{C}}$. Let $\boldsymbol{p}_{\boldsymbol{c}} = \boldsymbol{p}_{\boldsymbol{c}}(\boldsymbol{g},~\boldsymbol{c})$ and $\boldsymbol{O}_{\boldsymbol{c}} = \boldsymbol{O}_{\boldsymbol{c}}(\boldsymbol{g})$ denote the contact position and orientation obtained via forward kinematics, and define the grasp map
$\boldsymbol{G}_{\boldsymbol{c}} = \begin{bmatrix}\mathbf{I} \\ (\boldsymbol{p}_{\boldsymbol{c}} - \boldsymbol{m})_{\times}\end{bmatrix}\boldsymbol{O}_{\boldsymbol{c}}$.
Let $d_M(\cdot)$ be the signed distance to the object surface.
Given target wrenches $\{\boldsymbol{w}_j\}_{j=1}^J$, a scaling factor $\lambda > 0$, and weights $\kappa_\bullet > 0$, grasp synthesis for bimanual robots is formulated as:

\vspace{-2mm}
\begin{equation}
\begin{aligned}
\min_{\boldsymbol{g},\,\{\boldsymbol{f}_{\boldsymbol{c}}\}}\quad
& \kappa_{w} \sum_{j=1}^J \Big\| \lambda \boldsymbol{w}_j - \sum_{{\boldsymbol{c}}\in\mathcal{C}} \boldsymbol{G}_{\boldsymbol{c}}(\boldsymbol{g}) \boldsymbol{f}_{\boldsymbol{c}} \Big\|_2^2 \\
& \quad + \kappa_{\text{con}} \sum_{{\boldsymbol{c}}\in\mathcal{C}} \psi\!\big(d_M(\boldsymbol{p}_{\boldsymbol{c}})\big) \\
& \quad + \kappa_{\text{coll}} \,\Phi_{M}(\boldsymbol{g}) + \kappa_{\text{hh}} \,\Phi_{\text{hh}}(\boldsymbol{g})
\end{aligned}
\label{eq:main_obj}
\end{equation}

\vspace{-3mm}

\begin{align}
\text{s.t.}\quad
& \boldsymbol{q}_{h,\min} \le \boldsymbol{q}_h \le \boldsymbol{q}_{h,\max}, \quad h\in\{0,1\},
\label{eq:main_obj_q}\\
& \boldsymbol{f}_{\boldsymbol{c}} \in \mathcal{F}, \quad {\boldsymbol{c}} \in \mathcal{C},
\label{eq:main_obj_f}\\
& \boldsymbol{R}_h \in SO(3), \quad h\in\{0,1\},
\label{eq:main_obj_R}
\end{align}
where $\psi(d)$ is a distance energy that measures distance between contact points on the hands and object surface, $\Phi_{M}(\boldsymbol{g})$ is a hand--object collision energy based on signed-distance penalties, and $\Phi_{\text{hh}}(\boldsymbol{g})$ is an inter-hand penetration energy derived from hand--hand signed distances.

Following BODex~\cite{bodex}, we consider the above as a nonlinear bilevel program, where the lower level quadratic programming optimizes contact forces for each contact point to realize the target wrench and the upper level updates the hand pose to reduce the error between target and achievable wrenches via gradient descent. cuRobo~\cite{curobo} and a GPU-based QP solver are utilized to solve the problem efficiently.

\begin{figure}[ht]
    \centering
    \includegraphics[width=0.85\linewidth]{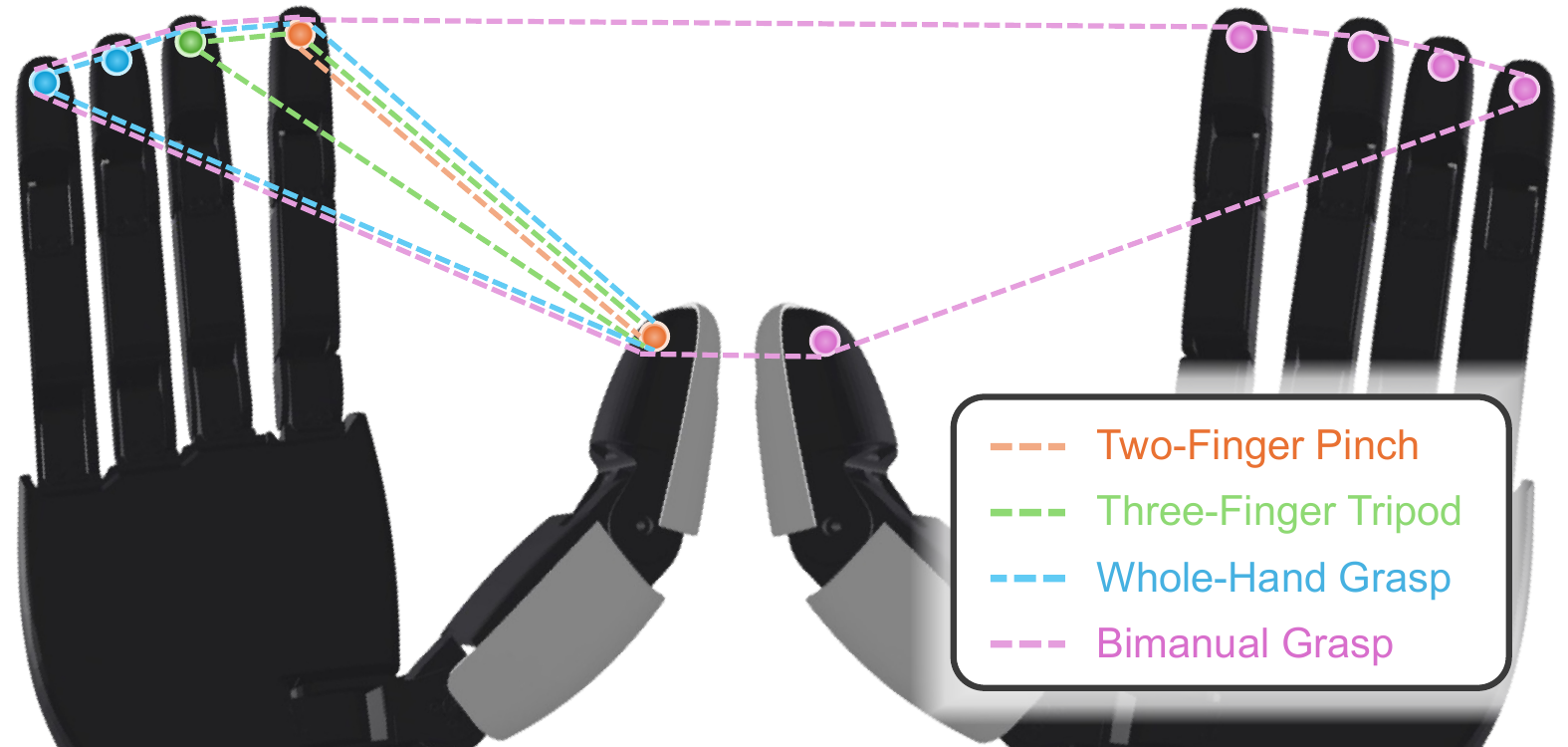}
    \caption{\textbf{Hand contact points for various grasp strategies.} Different grasp strategies select distinct fingertip contact points, which are used to compute energy terms in the optimization process of grasp synthesis.}
    \label{fig:hand_contact_point}
    \vspace{-3mm}
\end{figure}

The optimization program is unified for various grasp strategies. To adapt to a specific strategy, we select different contact points on the dexterous hands, as illustrated in Fig.~\ref{fig:hand_contact_point}. 
A gallery of synthesized grasps is presented in Fig.~\ref{fig:teaser}. 

For each object placed in the scene, we generate 500 candidate grasps. Through physical validation, we filter out physically implausible grasps. Next, we use cuRobo for inverse kinematics analysis to determine whether these grasps are reachable by bimanual robots, and perform collision checking to exclude grasps that result in collisions with objects other than the target. Finally, we calculate the $SE(3)$ distance between the poses of the remaining grasps and the current end-effector pose of the bimanual robots, ranking them accordingly and selecting the grasp with the shortest distance as the preferred grasp.
Specifically, the $SE(3)$ distance between two poses $\boldsymbol{T}_1$ and $\boldsymbol{T}_2$ is defined as:
\begin{equation}
d(\boldsymbol{T}_1, \boldsymbol{T}_2) = \|\boldsymbol{t}_1 - \boldsymbol{t}_2\|_2 + \lambda \cdot d_{\mathrm{rot}}(\boldsymbol{R}_1, \boldsymbol{R}_2),
\end{equation}
where $\boldsymbol{t}_1$, $\boldsymbol{t}_2$ are the translation vectors, $\boldsymbol{R}_1$, $\boldsymbol{R}_2$ are the rotation matrices, and $\lambda$ is a weighting factor. The rotation distance $d_{\mathrm{rot}}$ is given by:
\begin{equation}
d_{\mathrm{rot}}(\boldsymbol{R}_1, \boldsymbol{R}_2) = \arccos\left(\frac{\operatorname{trace}(\boldsymbol{R}_1^{-1} \boldsymbol{R}_2) - 1}{2}\right).
\end{equation}

This selection strategy favors grasps that require minimal motion from the current configuration, which not only improves the efficiency of the grasping process but also leads to more natural and smooth robot movements.

\subsection{Demonstration Generation}
\label{sec:demonstration_generation}

After obtaining the preferred grasp, we divide the entire grasping process into four stages: pregrasp---the end-effector moves to a position 0.1~m away from the generated grasp pose in the direction opposite to the palm to avoid unintended collisions during approach; grasp---the hand reaches the preferred grasp pose; squeeze---the fingers apply pressure to the object to achieve a stable grasp; and lift---the object is lifted by 0.2~m. We employ bimanual motion planning to generate collision-free coordinated trajectories. To prevent policies trained on such data from exhibiting hesitation, we merge adjacent steps with negligible movements. In simulation, the robot executes the planned trajectory and grasp using PD control, and we verify whether the object is stably lifted. Specifically, the object must be raised at least 0.17~m above its initial pose and remain elevated for at least one second without being dropped. 
If the object is successfully lifted and no unexpected contacts or movements are detected, the trajectory is recorded and rendered. By generating demonstrations across a diverse set of objects and various grasp strategies, we construct UltraDexGrasp-20M, a large-scale, multi-strategy grasp dataset for bimanual robots, comprising 20 million frames over 1,000 objects.

It is noteworthy that, similar to~\cite{dexpoint}, we supplement the robot's point cloud during the rendering process with an additional imaged point cloud.
During real-world deployment, the robot's joint positions are known, enabling us to also use simulation to generate an imaged point cloud.
This approach significantly reduces the sim-to-real gap.

\section{Universal Dexterous Grasp Policy}
\label{sec:policy}

\begin{figure*}[ht]
    \centering
    \includegraphics[width=0.98\linewidth]{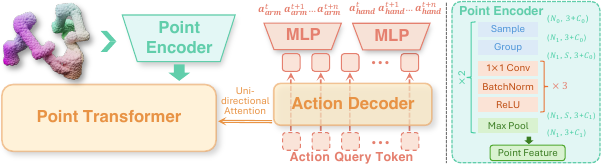}
    \vspace{1mm}
    \caption{\textbf{Overview of the policy architecture.} The proposed grasp policy takes point clouds as input, encodes them using a point encoder, aggregates scene features via unidirectional attention, and predicts control commands. The policy supports multiple grasp strategies and improves generalization across diverse objects.}
    \label{fig:policy}
    \vspace{-1mm}
\end{figure*}

\subsection{Overall Architecture}
\label{sec:overall_architecture}

To accommodate multiple grasp strategies and improve generalization across a wide range of objects, we propose a universal dexterous grasp policy, as illustrated in Fig.~\ref{fig:policy}. 
With simplicity and effectiveness in mind, we design the policy without redundant structures or auxiliary tasks.
The policy receives scene point clouds as input, which are encoded into point features using a point encoder. The point features are processed by a decoder-only transformer architecture. Finally, action decoders predict control commands. 

\subsection{Point Cloud Encoding}
\label{sec:point_cloud_encoding}

We first downsample the point cloud using farthest point sampling (FPS)~\cite{pointnet} to adjust the point cloud density. In practice, we use 2,048 points to balance computational cost and the granularity of scene representation. We then encode the resulting point cloud using a point encoder based on the PointNet++~\cite{pointnet++} architecture. Specifically, we employ two set abstraction layers. The first layer does not perform downsampling and thus maintains 2,048 points. For each point, a group is formed by identifying its 32 nearest neighbors (including the point itself) using k-nearest neighbors algorithm (k-NN). Local features for each group are extracted via a series of $1\times1$ convolution, batch normalization, ReLU activation function, and max pooling. The second set abstraction layer has a similar structure but downsamples the 2,048 points to 256, thereby capturing higher-level features for action prediction.

\subsection{Action Prediction}
\label{sec:action_prediction}

To read out action latents, a chunk of learnable action query tokens is fed into the transformer backbone, where they integrate scene information through unidirectional attention to the point features. Finally, a multi-layer perceptron (MLP) transforms the resulting action latents into action vectors to be executed by the robot.
Rather than directly regressing the action vectors, our decoder predicts a bounded Gaussian distribution over actions via a truncated normal parameterization and optimizes the negative log-likelihood of the ground truth actions. Modeling actions probabilistically yields more stable training and enhances overall performance.

\section{Experiments}
\label{sec:experiments}

We conduct comprehensive experiments in both simulation and real-world environments to evaluate the effectiveness of our data-generation pipeline for universal dexterous grasping, as well as the proposed universal dexterous grasp policy. Specifically, we aim to answer the following questions:

1) Does training on UltraDexGrasp-20M yield a policy with universal grasping capabilities and strong generalization across diverse objects?

2) How does the universal dexterous grasp policy perform compared to previous grasping methods and closed-loop control policies?

3) How does the policy perform as the amount of training data increases?

4) Do the key design components of the proposed policy effectively improve grasp success rates?

5) How does the policy trained on UltraDexGrasp-20M perform in real-world scenarios?

\subsection{Simulation Experiments}
\label{sec:exp_simulation}

\subsubsection{Experimental Setup}
We construct a dual-arm, dual-hand system composed of two 6-DoF UR5e robots and two 12-DoF XHand in simulation.  
The test set consists of 600 objects, including both seen and unseen categories during training. These objects exhibit significant variation in shape, weight, and size. The weight of the objects ranges from 5~g to 1,000~g. The smallest object has a bounding box whose longest edge is less than 0.03~m, while the largest object has a bounding box whose shortest edge is greater than 0.5~m.

\subsubsection{Baselines} 
To demonstrate the effectiveness of the proposed universal dexterous grasp policy, we select two strong baselines, DP3~\cite{dp3} and DexGraspNet~\cite{dexgraspnet}, for comparison. DP3 is a diffusion policy that takes point clouds and robot state as input and has shown strong performance in dexterous manipulation tasks. DexGraspNet takes the complete object mesh as input to generate grasp poses, and motion planning is utilized to obtain the full execution trajectory.

\subsubsection{Evaluation Metrics} 
Each policy is evaluated on 600 objects, with 10 trials conducted for each object. The objects are categorized into three groups based on size: small, medium, and large. 
For small and medium items, placements are randomized within a 0.8~m $\times$ 0.2~m area, while large items are randomly placed in a 0.15~m $\times$ 0.16~m region to ensure reachability for the robot’s end-effectors.
The success rate of each policy is reported.

\subsubsection{Main Results}
\begin{table}[ht]
    \centering
    \vspace{-1mm}
    \caption{\textbf{Results on simulation benchmarks.} The success rates (\%) of each policy are reported. The proposed grasp policy trained on UltraDexGrasp-20M demonstrates strong generalization to diverse objects and consistently outperforms the baselines. The best results are highlighted in \textbf{bold}.}
    \label{table:exp_sim_main}

    \begin{tabular}{l|c|c|c|c}
    \toprule
    Benchmark & Object Size & DP3 & DexGraspNet & \textbf{Ours} \\
    \midrule
    \multirow{3}{*}{Seen Objects} & Small & 41.7 & 45.6 & \textbf{78.8} \\
    & Medium & 54.3 & 72.0 & \textbf{84.3} \\
    & Large & 48.5 & - & \textbf{90.4} \\
    \midrule
    \multirow{3}{*}{Unseen Objects} & Small & 37.4 & 45.6 & \textbf{76.9} \\
    & Medium & 50.1 &  72.0 &\textbf{85.8} \\
    & Large & 48.1 & - &\textbf{87.5} \\
    \midrule
    \multicolumn{2}{c|}{\textbf{Average}} & 46.7 & 58.8 & \textbf{84.0} \\
    \bottomrule

    \end{tabular}

    \vspace{1mm}
\end{table}

To answer Questions 1 and 2, we conduct experiments on the simulation benchmark. Both DP3 and our policy are trained on the UltraDexGrasp-20M dataset, while DexGraspNet, an optimization-based method, does not require training. As shown in Table~\ref{table:exp_sim_main}, our policy achieves an average success rate of 84.0\%. It effectively handles small, medium, and large objects. 
Notably, it demonstrates strong generalization with an 83.4\% success rate on unseen objects. These results indicate that training on UltraDexGrasp-20M yields a policy with universal grasping capabilities and robust generalization across diverse objects. When trained on the same dataset, our policy outperforms DP3 by 37.3 percentage points, highlighting the superiority of the proposed grasp policy. 
We attribute this improvement to the carefully designed point encoder, which enables the policy to capture fine-grained object geometry, as well as the effective use of a unidirectional attention mechanism to aggregate scene features. 
DexGraspNet achieves an average success rate of 58.8\% on small and medium objects, with significantly lower performance on small objects compared to medium ones. Additionally, DexGraspNet, which can only synthesize unimanual grasps, cannot handle large objects. These findings underscore the importance of multi-strategy dexterous grasping for bimanual robots.

\subsubsection{Scaling with Training Data}
\begin{figure}[ht]
    \centering
    \vspace{-0.3mm}
    \includegraphics[width=0.85\linewidth]{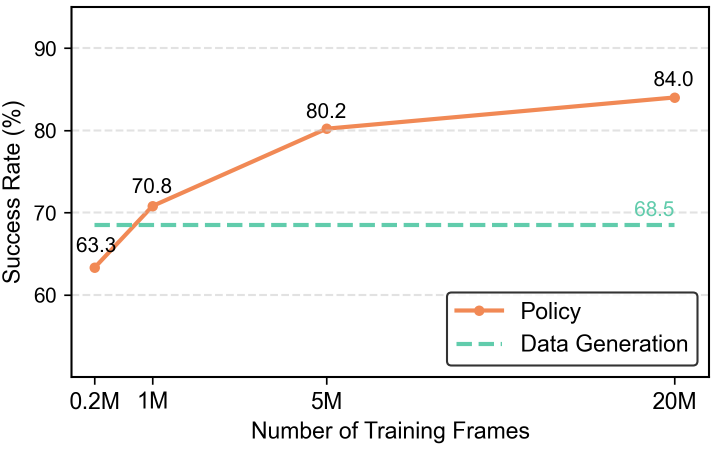}
    \vspace{-1mm}
    \caption{\textbf{The variation in performance with growing amounts of training data.} The performance of the policy consistently improves as the volume of data increases.}
    \label{fig:exp_data_scaling}
\end{figure}

To answer Question 3, we compare the average success rates of our policy across both seen and unseen objects, trained with different amounts of data. 
As shown in Fig.~\ref{fig:exp_data_scaling}, the overall performance of the policy consistently improves as the amount of training data increases.
For comparison, the average success rate of grasping data generation is 68.5\%. When training frames exceed 1M, the performance of the learned policy significantly surpasses that of data generation.

\subsubsection{Effectiveness of Design Choices}
\begin{table}[ht]
    \centering
    \caption{\textbf{Ablation study on policy design choices.} The success rates (\%) are reported. Both the bounded Gaussian distribution prediction and the unidirectional attention mechanism significantly improve performance.}
    \label{table:exp_design_choice}

    \begin{tabular}{c|c|c|c}
    \toprule
    & w/o Dist. Pred. & w/o Uni. Attn. &\textbf{Ours} \\
    \midrule
    \textbf{Success Rate (\%)} & 73.5 & 68.2 & \textbf{84.0} \\
    \bottomrule
    \end{tabular}

\end{table}

To answer Question 4, we conduct ablation studies on key policy design choices. Specifically, we compare our policy with two ablated versions: one without bounded Gaussian distribution prediction (w/o Dist. Pred.) and another without the unidirectional attention mechanism (w/o Uni. Attn.), where all tokens input to the transformer decoder can freely attend to each other. As reported in Tab.~\ref{table:exp_design_choice}, our policy achieves more than a 10\% improvement in success rate over both ablation baselines, demonstrating the effectiveness of our design choices.

\subsection{Real-World Experiments}
\label{sec:exp_real_world}

\begin{figure}[ht]
    \centering
    \includegraphics[width=0.98\linewidth]{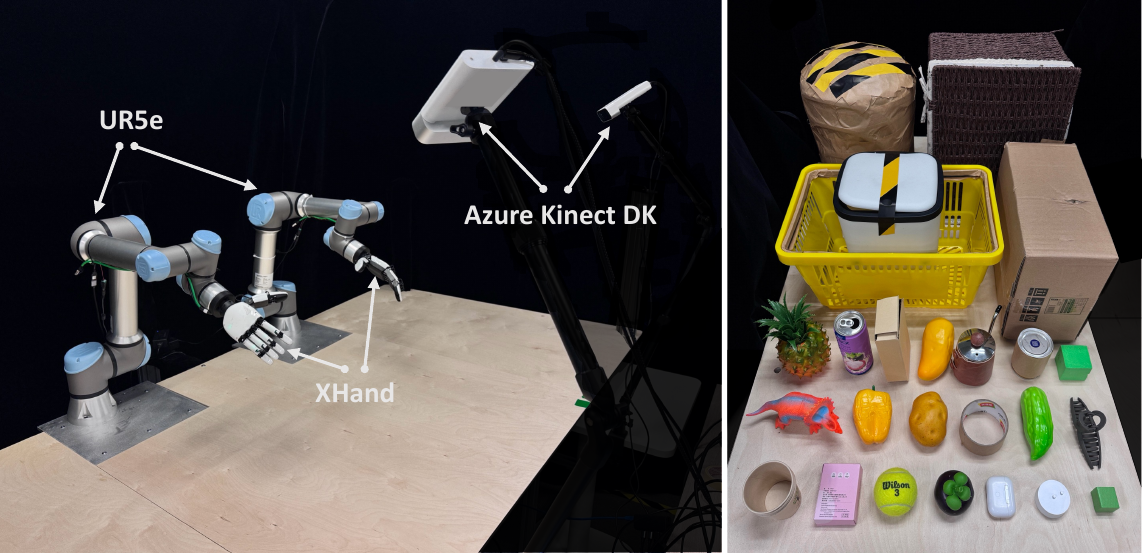}
    \caption{\textbf{Real-world experiment setup.} We employ two UR5e robots, two XHand, and two eye-on-base Azure Kinect DK cameras. A variety of objects are collected for testing.}
    \label{fig:real_setup}
\end{figure}

\subsubsection{Experimental Setup}
The real-world experimental setup is illustrated in Fig.~\ref{fig:real_setup}. Two UR5e robots are placed on a table at a distance of 0.9~{m} apart. Each robot’s end-effector is equipped with a 12-DoF XHand, attached via a flange. 
Two Azure Kinect DK cameras are positioned on the tabletop to capture scene point clouds as input for the policy. 
The control frequency of the robotic system is set to 10~Hz.

\subsubsection{Sim-to-Real Implementation Details}
To facilitate sim-to-real transfer, we implement several procedures. 
We establish a consistent coordinate system for both domains and calibrate the intrinsic and extrinsic parameters of the real-world cameras to obtain point clouds aligned with those in the simulation. Since depth cameras are subject to noise, Statistical Outlier Removal (SOR) is applied to filter out outlier points. 
Following~\cite{dexpoint}, imaged point clouds for robots are used to further mitigate the sim-to-real gap caused by low-quality, noisy, and incomplete point clouds obtained in the real world. 
Additionally, during demonstration generation, joint impedance randomization is employed to reduce the dynamics gap between the real world and simulation.

\subsubsection{Benchmarks}
We evaluate the grasping performance of the policies using 25 objects, which are categorized into small, medium, and large based on their sizes. Each policy is tested 15 times on each object, with a different object pose for each trial. Consistent with the simulation benchmark, objects in the small and medium categories are randomly placed within a 0.8~m $\times$ 0.2~m area, while large objects are randomly placed within a 0.15~m $\times$ 0.16~m area. 
DP3 and DexGraspNet are chosen as baselines. For DexGraspNet, which requires object poses and complete object meshes, object poses are estimated using FoundationPose~\cite{foundationpose}, and meshes are acquired through AR code scanning.  

\subsubsection{Main Results}
\begin{table}[ht]
    \centering
    \caption{\textbf{Results on real-world benchmarks.} The success rates (\%) are reported. The proposed grasp policy trained on UltraDexGrasp-20M demonstrates robust sim-to-real transfer and consistently outperforms the baselines. 
    }
    \label{table:exp_real_main}

    \begin{tabular}{c|c|c|c}
    \toprule
    Object Size & DP3 & DexGraspNet &\textbf{Ours} \\
    \midrule
    Small & 37.3 & 51.3 & \textbf{72.0} \\
    Medium & 56.0 & 73.3 &\textbf{82.2} \\
    Large & 46.7 & - &\textbf{89.3} \\
    \midrule
    {\textbf{Average}} & 46.7 & 62.3 & \textbf{81.2} \\
    \bottomrule
    \end{tabular}

\end{table}

To answer Question 5, we directly deploy the proposed universal dexterous grasp policy, trained on UltraDexGrasp-20M, in real-world scenarios. The policy demonstrates robust zero-shot sim-to-real transfer and successfully handles novel objects of various shapes, sizes, and weights. Among the tested objects, the smallest has a volume of only 18~$\mathrm{cm}^3$, while the largest is 26,400~$\mathrm{cm}^3$; the lightest weighs only 3.6~g, and the heaviest is 1,095~g. Our policy effectively adapts grasping strategies to these diverse objects, exhibiting strategies such as three-finger tripod, whole-hand grasp, and bimanual grasp. As presented in Tab.~\ref{table:exp_real_main}, our policy achieves robust grasps with an average success rate of 81.2\%, significantly outperforming the baselines.

\section{Conclusion}
\label{sec:conclusion_and_limitations}

In this paper, we introduce \textbf{UltraDexGrasp}, a framework for universal dexterous grasping with bimanual robots. We propose a data generation pipeline that features a complementary integration of optimization-based grasp synthesis and planning-based demonstration generation to produce universal dexterous grasp data for bimanual robots. With this pipeline, we curate \textbf{UltraDexGrasp-20M}, a large-scale multi-strategy grasp dataset, which contains 20 million frames across 1,000 objects.
Tested on 600 diverse objects in simulation, our proposed policy achieves an average success rate of 84.0\%.  
Deployed directly in the real world, the policy demonstrates robust zero-shot sim-to-real transfer.



\bibliographystyle{IEEEtran}
\bibliography{IEEEabrv}

@article{graspit,
  title={Graspit! a versatile simulator for robotic grasping},
  author={Miller, Andrew T and Allen, Peter K},
  journal={IEEE Robotics \& Automation Magazine},
  volume={11},
  number={4},
  pages={110--122},
  year={2004},
  publisher={IEEE}
}

@article{dexgraspnet,
  title={Dexgraspnet: A large-scale robotic dexterous grasp dataset for general objects based on simulation},
  author={Wang, Ruicheng and Zhang, Jialiang and Chen, Jiayi and Xu, Yinzhen and Li, Puhao and Liu, Tengyu and Wang, He},
  journal={arXiv preprint arXiv:2210.02697},
  year={2022}
}

@article{graspqp,
  title={GraspQP: Differentiable Optimization of Force Closure for Diverse and Robust Dexterous Grasping},
  author={Zurbr{\"u}gg, Ren{\'e} and Cramariuc, Andrei and Hutter, Marco},
  journal={arXiv preprint arXiv:2508.15002},
  year={2025}
}

@article{bodex,
  title={Bodex: Scalable and efficient robotic dexterous grasp synthesis using bilevel optimization},
  author={Chen, Jiayi and Ke, Yubin and Wang, He},
  journal={arXiv preprint arXiv:2412.16490},
  year={2024}
}

@article{liu2021synthesizing,
  title={Synthesizing diverse and physically stable grasps with arbitrary hand structures using differentiable force closure estimator},
  author={Liu, Tengyu and Liu, Zeyu and Jiao, Ziyuan and Zhu, Yixin and Zhu, Song-Chun},
  journal={IEEE Robotics and Automation Letters},
  volume={7},
  number={1},
  pages={470--477},
  year={2021},
  publisher={IEEE}
}

@article{li2022gendexgrasp,
  title={Gendexgrasp: Generalizable dexterous grasping},
  author={Li, Puhao and Liu, Tengyu and Li, Yuyang and Geng, Yiran and Zhu, Yixin and Yang, Yaodong and Huang, Siyuan},
  journal={arXiv preprint arXiv:2210.00722},
  year={2022}
}

@article{dex1b,
  title={Dex1B: Learning with 1B Demonstrations for Dexterous Manipulation},
  author={Ye, Jianglong and Wang, Keyi and Yuan, Chengjing and Yang, Ruihan and Li, Yiquan and Zhu, Jiyue and Qin, Yuzhe and Zou, Xueyan and Wang, Xiaolong},
  journal={arXiv preprint arXiv:2506.17198},
  year={2025}
}

@inproceedings{goldfeder2009columbia,
  title={The columbia grasp database},
  author={Goldfeder, Corey and Ciocarlie, Matei and Dang, Hao and Allen, Peter K},
  booktitle={2009 IEEE international conference on robotics and automation},
  pages={1710--1716},
  year={2009},
  organization={IEEE}
}

@inproceedings{unidexgrasp,
  title={Unidexgrasp: Universal robotic dexterous grasping via learning diverse proposal generation and goal-conditioned policy},
  author={Xu, Yinzhen and Wan, Weikang and Zhang, Jialiang and Liu, Haoran and Shan, Zikang and Shen, Hao and Wang, Ruicheng and Geng, Haoran and Weng, Yijia and Chen, Jiayi and others},
  booktitle={Proceedings of the IEEE/CVF Conference on Computer Vision and Pattern Recognition},
  pages={4737--4746},
  year={2023}
}

@inproceedings{unidexgrasp++,
  title={Unidexgrasp++: Improving dexterous grasping policy learning via geometry-aware curriculum and iterative generalist-specialist learning},
  author={Wan, Weikang and Geng, Haoran and Liu, Yun and Shan, Zikang and Yang, Yaodong and Yi, Li and Wang, He},
  booktitle={Proceedings of the IEEE/CVF International Conference on Computer Vision},
  pages={3891--3902},
  year={2023}
}

@article{dextrahg,
  title={Dextrah-g: Pixels-to-action dexterous arm-hand grasping with geometric fabrics},
  author={Lum, Tyler Ga Wei and Matak, Martin and Makoviychuk, Viktor and Handa, Ankur and Allshire, Arthur and Hermans, Tucker and Ratliff, Nathan D and Van Wyk, Karl},
  journal={arXiv preprint arXiv:2407.02274},
  year={2024}
}

@article{dextrahrgb,
  title={Dextrah-rgb: Visuomotor policies to grasp anything with dexterous hands},
  author={Singh, Ritvik and Allshire, Arthur and Handa, Ankur and Ratliff, Nathan and Van Wyk, Karl},
  journal={arXiv preprint arXiv:2412.01791},
  year={2024}
}

@article{wei2022dvgg,
  title={DVGG: Deep variational grasp generation for dextrous manipulation},
  author={Wei, Wei and Li, Daheng and Wang, Peng and Li, Yiming and Li, Wanyi and Luo, Yongkang and Zhong, Jun},
  journal={IEEE Robotics and Automation Letters},
  volume={7},
  number={2},
  pages={1659--1666},
  year={2022},
  publisher={IEEE}
}

@article{realdex,
  title={Realdex: Towards human-like grasping for robotic dexterous hand},
  author={Liu, Yumeng and Yang, Yaxun and Wang, Youzhuo and Wu, Xiaofei and Wang, Jiamin and Yao, Yichen and Schwertfeger, S{\"o}ren and Yang, Sibei and others},
  journal={arXiv preprint arXiv:2402.13853},
  year={2024}
}

@inproceedings{contactpose,
  title={ContactPose: A dataset of grasps with object contact and hand pose},
  author={Brahmbhatt, Samarth and Tang, Chengcheng and Twigg, Christopher D and Kemp, Charles C and Hays, James},
  booktitle={European Conference on Computer Vision},
  pages={361--378},
  year={2020},
}

@inproceedings{dexycb,
  title={DexYCB: A benchmark for capturing hand grasping of objects},
  author={Chao, Yu-Wei and Yang, Wei and Xiang, Yu and Molchanov, Pavlo and Handa, Ankur and Tremblay, Jonathan and Narang, Yashraj S and Van Wyk, Karl and others},
  booktitle={Proceedings of the IEEE/CVF conference on computer vision and pattern recognition},
  pages={9044--9053},
  year={2021}
}

@inproceedings{li2023frogger,
  title={Frogger: Fast robust grasp generation via the min-weight metric},
  author={Li, Albert H and Culbertson, Preston and Burdick, Joel W and Ames, Aaron D},
  booktitle={2023 IEEE/RSJ International Conference on Intelligent Robots and Systems (IROS)},
  pages={6809--6816},
  year={2023},
  organization={IEEE}
}

@inproceedings{turpin2022grasp,
  title={Grasp’d: Differentiable contact-rich grasp synthesis for multi-fingered hands},
  author={Turpin, Dylan and Wang, Liquan and Heiden, Eric and Chen, Yun-Chun and Macklin, Miles and Tsogkas, Stavros and Dickinson, Sven and Garg, Animesh},
  booktitle={European Conference on Computer Vision},
  pages={201--221},
  year={2022},
  organization={Springer}
}

@article{turpin2023fast,
  title={Fast-grasp'd: Dexterous multi-finger grasp generation through differentiable simulation},
  author={Turpin, Dylan and Zhong, Tao and Zhang, Shutong and Zhu, Guanglei and Liu, Jingzhou and Singh, Ritvik and Heiden, Eric and Macklin, Miles and Tsogkas, Stavros and Dickinson, Sven and others},
  journal={arXiv preprint arXiv:2306.08132},
  year={2023}
}

@article{bimangrasp,
  title={Bimanual grasp synthesis for dexterous robot hands},
  author={Shao, Yanming and Xiao, Chenxi},
  journal={IEEE Robotics and Automation Letters},
  year={2024},
  publisher={IEEE}
}

@inproceedings{dexpoint,
  title={Dexpoint: Generalizable point cloud reinforcement learning for sim-to-real dexterous manipulation},
  author={Qin, Yuzhe and Huang, Binghao and Yin, Zhao-Heng and Su, Hao and Wang, Xiaolong},
  booktitle={Conference on Robot Learning},
  pages={594--605},
  year={2023},
  organization={PMLR}
}

@article{lin2025sim,
  title={Sim-to-real reinforcement learning for vision-based dexterous manipulation on humanoids},
  author={Lin, Toru and Sachdev, Kartik and Fan, Linxi and Malik, Jitendra and Zhu, Yuke},
  journal={arXiv preprint arXiv:2502.20396},
  year={2025}
}

@article{graspvla,
  title={Graspvla: a grasping foundation model pre-trained on billion-scale synthetic action data},
  author={Deng, Shengliang and Yan, Mi and Wei, Songlin and Ma, Haixin and Yang, Yuxin and Chen, Jiayi and Zhang, Zhiqi and Yang, Taoyu and Zhang, Xuheng and Cui, Heming and others},
  journal={arXiv preprint arXiv:2505.03233},
  year={2025}
}

@article{dexvlg,
  title={DexVLG: Dexterous Vision-Language-Grasp Model at Scale},
  author={He, Jiawei and Li, Danshi and Yu, Xinqiang and Qi, Zekun and Zhang, Wenyao and Chen, Jiayi and Zhang, Zhaoxiang and Zhang, Zhizheng and Yi, Li and Wang, He},
  journal={arXiv preprint arXiv:2507.02747},
  year={2025}
}

@inproceedings{dexgraspnet2.0,
  title={Dexgraspnet 2.0: Learning generative dexterous grasping in large-scale synthetic cluttered scenes},
  author={Zhang, Jialiang and Liu, Haoran and Li, Danshi and Yu, XinQiang and Geng, Haoran and Ding, Yufei and Chen, Jiayi and Wang, He},
  booktitle={8th Annual Conference on Robot Learning},
  year={2024}
}

@article{anygrasp,
  title={Anygrasp: Robust and efficient grasp perception in spatial and temporal domains},
  author={Fang, Hao-Shu and Wang, Chenxi and Fang, Hongjie and Gou, Minghao and Liu, Jirong and Yan, Hengxu and Liu, Wenhai and Xie, Yichen and Lu, Cewu},
  journal={IEEE Transactions on Robotics},
  volume={39},
  number={5},
  pages={3929--3945},
  year={2023},
  publisher={IEEE}
}

@inproceedings{6dofgraspnet,
  title={6-dof graspnet: Variational grasp generation for object manipulation},
  author={Mousavian, Arsalan and Eppner, Clemens and Fox, Dieter},
  booktitle={Proceedings of the IEEE/CVF international conference on computer vision},
  pages={2901--2910},
  year={2019}
}

@inproceedings{graspnet,
  title={Graspnet-1billion: A large-scale benchmark for general object grasping},
  author={Fang, Hao-Shu and Wang, Chenxi and Gou, Minghao and Lu, Cewu},
  booktitle={Proceedings of the IEEE/CVF conference on computer vision and pattern recognition},
  pages={11444--11453},
  year={2020}
}

@article{dexgraspvla,
  title={Dexgraspvla: A vision-language-action framework towards general dexterous grasping},
  author={Zhong, Yifan and Huang, Xuchuan and Li, Ruochong and Zhang, Ceyao and Liang, Yitao and Yang, Yaodong and Chen, Yuanpei},
  journal={arXiv preprint arXiv:2502.20900},
  year={2025}
}

@inproceedings{dexgraspanything,
  title={Dexgrasp anything: Towards universal robotic dexterous grasping with physics awareness},
  author={Zhong, Yiming and Jiang, Qi and Yu, Jingyi and Ma, Yuexin},
  booktitle={Proceedings of the Computer Vision and Pattern Recognition Conference},
  pages={22584--22594},
  year={2025}
}

@article{fungrasp,
  title={FunGrasp: functional grasping for diverse dexterous hands},
  author={Huang, Linyi and Zhang, Hui and Wu, Zijian and Christen, Sammy and Song, Jie},
  journal={IEEE Robotics and Automation Letters},
  year={2025},
  publisher={IEEE}
}

@article{getagrip,
  title={Get a grip: Multi-finger grasp evaluation at scale enables robust sim-to-real transfer},
  author={Lum, Tyler Ga Wei and Li, Albert H and Culbertson, Preston and Srinivasan, Krishnan and Ames, Aaron D and Schwager, Mac and Bohg, Jeannette},
  journal={arXiv preprint arXiv:2410.23701},
  year={2024}
}

@article{robustdexgrasp,
  title={RobustDexGrasp: Robust Dexterous Grasping of General Objects},
  author={Zhang, Hui and Wu, Zijian and Huang, Linyi and Christen, Sammy and Song, Jie},
  journal={arXiv preprint arXiv:2504.05287},
  year={2025}
}

@article{clutterdexgrasp,
  title={ClutterDexGrasp: A Sim-to-Real System for General Dexterous Grasping in Cluttered Scenes},
  author={Chen, Zeyuan and Yan, Qiyang and Chen, Yuanpei and Wu, Tianhao and Zhang, Jiyao and Ding, Zihan and Li, Jinzhou and Yang, Yaodong and Dong, Hao},
  journal={arXiv preprint arXiv:2506.14317},
  year={2025}
}

@article{pointnet++,
  title={Pointnet++: Deep hierarchical feature learning on point sets in a metric space},
  author={Qi, Charles Ruizhongtai and Yi, Li and Su, Hao and Guibas, Leonidas J},
  journal={Advances in neural information processing systems},
  volume={30},
  year={2017}
}

@misc{curobo,
      title={cuRobo: Parallelized Collision-Free Minimum-Jerk Robot Motion Generation},
      author={Balakumar Sundaralingam and Siva Kumar Sastry Hari and Adam Fishman and Caelan Garrett
              and Karl Van Wyk and Valts Blukis and Alexander Millane and Helen Oleynikova and Ankur Handa
              and Fabio Ramos and Nathan Ratliff and Dieter Fox},
      year={2023},
      eprint={2310.17274},
      archivePrefix={arXiv},
      primaryClass={cs.RO}
}

@inproceedings{dp3,
	title={3D Diffusion Policy: Generalizable Visuomotor Policy Learning via Simple 3D Representations},
	author={Yanjie Ze and Gu Zhang and Kangning Zhang and Chenyuan Hu and Muhan Wang and Huazhe Xu},
	booktitle={Proceedings of Robotics: Science and Systems (RSS)},
	year={2024}
}

@inproceedings{foundationpose,
  title={Foundationpose: Unified 6d pose estimation and tracking of novel objects},
  author={Wen, Bowen and Yang, Wei and Kautz, Jan and Birchfield, Stan},
  booktitle={Proceedings of the IEEE/CVF Conference on Computer Vision and Pattern Recognition},
  pages={17868--17879},
  year={2024}
}

@article{anydexgrasp,
  title={AnyDexGrasp: General Dexterous Grasping for Different Hands with Human-level Learning Efficiency},
  author={Fang, Hao-Shu and Yan, Hengxu and Tang, Zhenyu and Fang, Hongjie and Wang, Chenxi and Lu, Cewu},
  journal={arXiv preprint arXiv:2502.16420},
  year={2025}
}

@inproceedings{jiang2021hand,
  title={Hand-object contact consistency reasoning for human grasps generation},
  author={Jiang, Hanwen and Liu, Shaowei and Wang, Jiashun and Wang, Xiaolong},
  booktitle={Proceedings of the IEEE/CVF international conference on computer vision},
  pages={11107--11116},
  year={2021}
}

@inproceedings{pointnet,
  title={Pointnet: Deep learning on point sets for 3d classification and segmentation},
  author={Qi, Charles R and Su, Hao and Mo, Kaichun and Guibas, Leonidas J},
  booktitle={Proceedings of the IEEE conference on computer vision and pattern recognition},
  pages={652--660},
  year={2017}
}

@article{drograsp,
  title={D (r, o) grasp: A unified representation of robot and object interaction for cross-embodiment dexterous grasping},
  author={Wei, Zhenyu and Xu, Zhixuan and Guo, Jingxiang and Hou, Yiwen and Gao, Chongkai and Cai, Zhehao and Luo, Jiayu and Shao, Lin},
  journal={arXiv preprint arXiv:2410.01702},
  year={2024}
}

@inproceedings{dexhanddiff,
  title={Interaction-aware diffusion planning for adaptive dexterous manipulation},
  author={Liang, Zhixuan and Mu, Yao and Wang, Yixiao and Chen, Tianxing and Shao, Wenqi and Zhan, Wei and Tomizuka, Masayoshi and Luo, Ping and Ding, Mingyu},
  booktitle={Proceedings of the Computer Vision and Pattern Recognition Conference},
  pages={1745--1755},
  year={2025}
}

\end{document}